\documentclass[11pt,a4paper]{article}
\usepackage[hyperref]{acl2017}
\usepackage{times}
\usepackage{latexsym}

\usepackage{url}
\usepackage{multirow}
\usepackage{amsmath}
\usepackage{tikz}

\usepackage{algorithm}
\usepackage{algorithmic}

\usepackage{subfig}

\aclfinalcopy 

\newcommand*{\affaddr}[1]{#1}
\newcommand*{\affmark}[1][*]{\textsuperscript{#1}}

\newcommand*{\email}[1]{\texttt{#1}}

\title{Top-Rank Enhanced Listwise Optimization \\ for Statistical Machine Translation}

\newlength{\astwd}
\author{%
Huadong Chen,\affmark[\dag] Shujian Huang,\affmark[\dag]\thanks{\hspace{\astwd}Corresponding author.}\hspace{\astwd} David Chiang,\affmark[\ddag] Xinyu Dai,\affmark[\dag] Jiajun Chen\affmark[\dag]\\
\affaddr{\affmark[\dag]State Key Laboratory for Novel Software Technology, Nanjing University}\\
\email{\{chenhd,huangsj,daixinyu,chenjj\}@nlp.nju.edu.cn}\\
\affaddr{\affmark[\ddag]Department of Computer Science and Engineering, University of Notre Dame}\\
\email{ dchiang@nd.edu}\\
}

\date{}

\begin{document}
\maketitle
\begin{abstract}
  Pairwise ranking methods are the basis of many widely used discriminative training approaches for structure prediction problems in natural language processing (NLP). Decomposing the problem of ranking hypotheses into pairwise comparisons enables simple and efficient solutions. However, neglecting the global ordering of the hypothesis list may hinder learning. We propose a listwise learning framework for structure prediction problems such as machine translation. Our framework directly models the entire translation list's ordering to learn parameters which may better fit the given listwise samples. Furthermore, we propose \emph{top-rank enhanced} loss functions, which are more sensitive to ranking errors at higher positions. Experiments on a large-scale Chinese-English translation task show that both our listwise learning framework and top-rank enhanced listwise losses lead to significant improvements in translation quality.
\end{abstract}

\section{Introduction}

Discriminative training methods for structured prediction in natural language processing (NLP) aim to estimate the parameters of a model that assigns a score to each hypothesis in the (possibly very large) search space.
For example, in statistical machine translation (SMT), the model assigns a score to each possible translation, and in syntactic parsing, the function assigns a score to each possible syntactic tree.
Ideally, the model should assign scores that rank hypotheses according to their true quality.
In this paper, we consider the problem of discriminative training for SMT.

Traditional SMT systems use log-linear models with only about a dozen features, such as translation probabilities and language model probabilities~\cite{Yamada2001,Koehn2003,Chiang2005,Liu2006}. These models can be tuned by minimum error rate training (MERT)~\cite{Och2003}, which directly optimizes BLEU using coordinate ascent combined with a global line search.

To enable training of modern SMT systems, which can have thousands of features or more, many research efforts have been made towards scalable discriminative training methods~\cite{Chiang2008,Hopkins:2011,Bazrafshan:2012}.
Most of these methods either define loss functions that push the model to correctly compare pairs of hypotheses, or use approximate optimization methods that effectively do the same.
For practical reasons, only a subset of the pairs are considered; these pairs are selected by either sampling~\cite{Hopkins:2011} or heuristic methods~\cite{Watanabe2007,Chiang2008}.

But this pairwise approach neglects the global ordering of the list of hypotheses, which may lead to problems trying to learn good parameter values.
Inspired by research in information retrieval (IR)~\cite{Cao2007,Xia2008}, we propose to directly model the ordering of the whole translation list, instead of decomposing it into translation pairs.

Previous research has tried to integrate listwise methods into SMT, but almost all of them focus on the reranking task, which aims to rescore the fixed translation lists generated by a baseline system. They try to either use listwise approaches to training the reranking model~\cite{Li2013,Niehues2015} or replace the pointwise ranking function, i.e. the log-linear model, with a listwise ranking function by introducing listwise features~\cite{zhang2016}.
In this paper, we focus on listwise approaches that can learn better discriminative models for SMT. We present a listwise learning framework for tuning translation systems that uses two listwise ranking objectives originally developed for IR, ListNet~\cite{Cao2007} and ListMLE~\cite{Xia2008}. But unlike standard IR problems, structured prediction problems usually have a huge search space, and at each training iteration, the list of search results can vary. The usual strategy is to form the union of all lists of search results, but this can lead to a ``patchy'' list that doesn't represent the full search space well. The listwise approaches always based on the permutation probability distribution over the list. Modeling the distribution over a ``patchy'' list, whose elements were generated by different parameters will affect listwise approaches' performance. To address this issue, we design an \emph{instance-aggregating} method: Instead of treating the data as a fixed-size set of lists that each grow over time as new translations are added at each iteration, we treat the data as a growing set of lists; each time a sentence is translated, the $k$-best list of translations is added as a new list.

We also extend standard listwise training by considering the importance of different instances in the list. Based on the intuition that instances at the top may be more important for ranking, we propose \emph{top-rank enhanced} loss functions, which incorporate a position-dependent cost that penalizes errors occurring at the top of the list more strongly.

We conduct large-scale Chinese-to-English translation experiments 
showing that our top-rank enhanced listwise learning methods significantly outperform other tuning methods with high dimensional feature sets. Additionally, even with a small basic feature set, our methods still obtain better results than MERT.

\section{Background}

\subsection{Log-linear models}

In this paper, we assume a log-linear model, which defines a scoring function on target translation hypotheses $\mathbf{e}$, given a source sentence $\mathbf{f}$:
\begin{align}\label{eq-log-linear}
Pr(\mathbf{e}\mid\mathbf{f}) &=\frac    {\exp s(\mathbf{e},\mathbf{f})}{\sum_{\mathbf{e'}} \exp s(\mathbf{e'},\mathbf{f})} \\
s(\mathbf{e},\mathbf{f}) &= \mathbf{w} \cdot \mathbf{h(\mathbf{e}\mid\mathbf{f})}
\end{align}
where $\mathbf{h}(\mathbf{e}\mid\mathbf{f})$ is the feature vector and $\mathbf{w}$ is the feature weight vector.

The process of training a SMT system includes both learning the sub-models, which are included in the feature vector $\mathbf{h}$, and learning the weight vector $\mathbf{w}$.

Then the decoding of SMT systems can be formulated as a search for the translation $\mathbf{\hat{e}}$ with the highest model score:

\begin{equation}\label{eq-decoding}
\mathbf{\hat{e}} = \arg\max_{\mathbf{e}\in\mathcal{E}}{s(\mathbf{e}, \mathbf{f})}
\end{equation}
where $\mathcal{E}$ is the set of all reachable hypotheses.

\subsection{SMT Features}\label{subsec-features}
In this paper, we use a hierarchical phrase based translation system~\cite{Chiang2005}. For convenient comparison, we divide features of SMT into the following three sets.

\textbf{Basic Features:} The basic features are those commonly used in hierarchical phrase based translation systems, including a language model, four translation model features, word, phrase and rule penalties, and penalties for unknown words, the glue rule and null translations.

\textbf{Extended Features:} Inspired by \newcite{Chen2013}, we manually group the parallel training data into 15 sets, according to their genre and origin. The translation models trained on each set are used as separate features. We also add an indicator feature for each individual training set to mark where the translation rule comes from. The extended features provide additional 60 translation model features and 16 indicator features, which is too many to be tuned with MERT.

\begin{figure}
	\centering
	\includegraphics[width=0.3\textwidth]{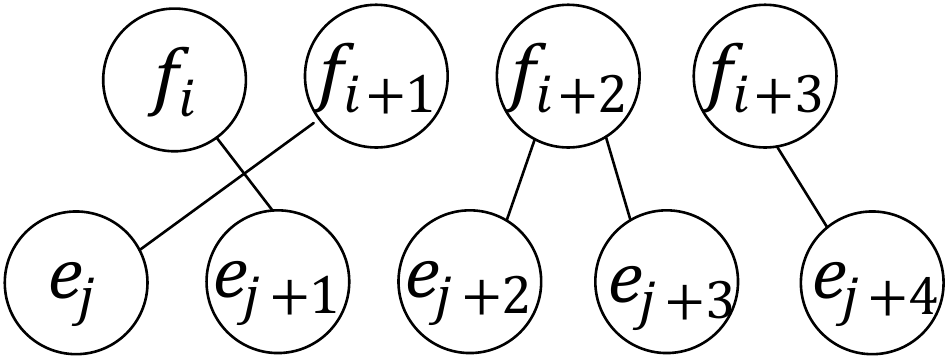}
	\caption{An example of word-phrase features for a phrase translation. The $f_i$ and $e_j$ represent the $i$-th in the source phrase and $j$-th word in the target phrase, respectively. }
	\label{fig-pairs}
\end{figure}

\textbf{Sparse Features:}  We use word-phrase pair features as our sparse features, which reflect the word-phrase correspondence in a hierarchical phrase~\cite{Watanabe2007}. Figure \ref{fig-pairs} illustrates an example of word-phrase pair features for a phrase translation pair $f_i, ..., f_{i+3}$ and $e_j, ..., e _{j+4}$. Word-phrase pair features $(f_i, e_{j+1})$, $(f_{i+1}, e_j)$, $(f_{i+2},e_{j+2}e_{j+3})$, $(f_{i+3}, e_{j+4})$ will be fired for the translation rule with the given word alignment. In practice, these feature only fire when all the source and target words in the feature are both in the top 100 most frequent words.

\subsection{Tuning via Pairwise Ranking}
The beam search strategy for SMT decoding process makes it convenient to get a $k$-best translation list for each source sentence. Given a set of source sentences and their corresponding translation lists, the tuning problem could be regarded as a ranking task. Many recently proposed SMT tuning methods are based on the pairwise ranking framework~\cite{Chiang2008,Hopkins:2011,Bazrafshan:2012}.

Pairwise ranking optimization (PRO)~\cite{Hopkins:2011} is a commonly used tuning method. The idea of PRO is to sample pairs $(\mathbf{e},\mathbf{e'})$ from the $k$-best list, and train a linear binary classifier to predict whether $\textit{eval}(\mathbf{e}) > \textit{eval}(\mathbf{e'})$ or $\textit{eval}(\mathbf{e}) < \textit{eval}(\mathbf{e'})$, where $\textit{eval}(\mathord\cdot)$ is an extrinsic metric like BLEU. In this paper, we use sentence-level BLEU with add-one smoothing \cite{lin+och:2004}.

The method gets a comparable BLEU score to MERT and MIRA~\cite{Chiang2008}, and scales well on large feature sets. Other pairwise ranking methods employ similar procedures.

\section{Listwise Learning Framework}\label{sec-ltuning}
Although ranking methods have shown their effectiveness in tuning for SMT systems~\cite{Hopkins:2011,Watanabe2012,Dreyer2015}, most proposed ranking approaches view tuning as pairwise ranking. These approaches decompose the ranking of the hypothesis list into pairs, which might limit the training method's ability to learn better parameters. To preserve the ranking information, we first formulate training as an instance of the listwise ranking problem. Then we propose a learning method based on the iterative learning framework of SMT tuning and further investigate the top-rank enhanced losses.

\subsection{Training Objectives}\label{ssec-objs}

\subsubsection{The Permutation Probability Model}

In order to directly model the translation list, we first introduce a probabilistic model proposed by~\newcite{Guiver2009}.
A ranking of a list of $k$ translations can be thought of as a function $\pi$ from $[1,k]$ to translations, where each $\pi(t)$ is the $t$-th translation candidate in the ranking. A scoring function $z$ (which could be either the model score, $s$, or the BLEU score, $\textit{eval}$) induces a probability distribution over rankings:
\begin{equation}\label{eq-rankingprob}
P_{z}(\pi) = \prod_{j =1}^{k}{\frac {\exp z({\pi(j)})} {\sum_{t = j}^{k}\exp z({\pi(t)})}}.
\end{equation}

\subsubsection{Loss Functions}

Based on the probabilistic model above, the loss function can be defined as the difference between the distribution over the ranking according to $eval(\cdot)$ and $s(\cdot)$. Thus, we introduce the following two standard listwise losses.

{\bf ListNet:} The ListNet loss is the cross entropy between the distributions calculated from $eval(\cdot)$ and $s(\cdot)$, respectively, over all permutations.

Due to the exponential number of permutations, \newcite{Cao2007} propose a top-one loss instead. Given the function $eval(\cdot)$ and $s(\cdot)$, the top-one loss is defined as:
\begin{align*}
L_{\text{Net-T}} &=  - \sum_{j=1}^{k}{P'_{eval}(\mathbf{e}_{j})\ \log P'_{s}(\mathbf{e}_{j})}\\
P'_z(\mathbf{e}_j)
&=\frac{\exp z(\mathbf{e}_j)}{\sum_{i=1}^k{\exp z(\mathbf{e}_i)}}
\end{align*}
where $\mathbf{e}_j$ is the $j$-th element in the $k$-best list, and $P'_z(\mathbf{e}_j)$ is the probability that translation $\mathbf{e}_j$ is ranked at the top by the function $z$.

{\bf ListMLE:} The ListMLE loss is the negative log-likelihood of the permutation probability of the correct ranking $\pi_{eval}$, calculated according to~$s(\cdot)$~\cite{Xia2008}:
\begin{equation}\label{eq-lossmle}
\begin{split}
L_{\text{MLE}} &=-\log P_s(\pi_{eval})\\
&= -\sum_{j = 1}^{k} \log \frac {\exp s({\pi_{eval}(j)})} {\sum_{t = j}^{k}\exp s({\pi_{eval}(t)})}.
\end{split}
\end{equation}

The training objective, which we want to minimize, is simply the total loss over all the lists in the tuning set.

\subsection{Training with Instance Aggregating}\label{ssec-procedure}

Because there can be exponentially many possible translations of a sentence, it's only feasible to rank the $k$ best translations rather than all of them; because the feature weights change at each iteration, we have a different $k$-best list to rank at each iteration. This is different from standard ranking problems in which the training instances stay the same each iteration. 

\begin{algorithm}
	\caption{MERT-like tuning algorithm}
	\label{alg-arc}
	\begin{algorithmic}[1]
		\REQUIRE  Training sentences $\{\mathbf{f}\}$, maximum number of iterations $I$, randomly initialized model parameters $\mathbf{w}^0$.
		\FOR{$i=0$ to $I$}
		  \FOR {source sentences $\mathbf{f}$}
		\STATE Decode $\mathbf{f}$:
		$\mathcal{E}_{\mathbf{f}}^{i} = \text{KbestDecoder}(\mathbf{f}, \mathbf{w}^i)$
        \STATE $T \leftarrow T \cup \{ \mathcal{E}_{\mathbf{f}}^i \}$
		\ENDFOR
		\STATE Training: $\mathbf{w}^{i+1} = \text{Optimization}(T, \mathbf{w}^{i})$
		\ENDFOR
	\end{algorithmic}

\end{algorithm}

Many previous tuning methods address this problem by merging the $k$-best list at the current iteration with the $k$-best lists at all previous iterations into a single list~\cite{Hopkins:2011}. We call this \emph{$k$-best merging}.
More formally, if $\mathcal{E}_{\mathbf{f}}^i$ is the $k$-best list of source sentence $\mathbf{f}$ at iteration $i$, then at each iteration, the model is trained on the set of lists:
\begin{align*}
\mathcal{E}_{\mathbf{f}} &= \bigcup_{j=0}^i \mathcal{E}_{\mathbf{f}}^j \\
T &= \left\{ \mathcal{E}_{\mathbf{f}} \mid \forall \mathbf{f} \right\}
\end{align*}
For each source sentence $\mathbf{f}$, $T$ has only one training sample, which is a better and better approximation to the full hypothesis set of $\mathbf{f}$ as more iterations pass.

Unlike previous tuning methods, our tuning method focuses on the distribution over permutation of the whole list. Moreover, unlike with listwise optimization methods used in IR, the $k$-best list produced for a source sentence at one iteration can differ dramatically from the $k$-best list produced at the next iteration. Merging $k$-best lists across iterations, each of which represents only a tiny fraction of the full search space, will lead to a ``patchy'' list that may hurt the learning performance of the listwise optimization algorithms.

To address this challenge, we propose \emph{instance aggregating}: instead of merging $k$-best lists across different iterations, we view the translation lists from different iterations as individual training instances:
\begin{align*}
T &= \{ \mathcal{E}_{\mathbf{f}}^j \mid \forall \mathbf{f}, 0 \leq j \leq i \}.
\end{align*}
With this method, each source sentence $\mathbf{f}$ has $i$ training instances at the $i$-th training iteration. In this way, we avoid ``patchy'' lists and obtain a better set of instances for tuning.

\begin{algorithm}[t]
	\caption{Listwise Optimization Algorithm}
	\label{alg-training}
	\begin{algorithmic}[1]
		\REQUIRE  Training instances $T$, model parameters $\mathbf{w}$, maximum number of epochs $J$, batch size $b$, number of batches $B$
		\FOR{ $j=0$ to $J$}
		\FOR{ $i=0$ to $B$}
		\STATE Sample a minibatch of $b$ lists from $T$ without replacement
		\STATE Calculate loss function $L$
		\STATE Calculate gradient $\nabla L$
		\STATE $\mathbf{w}_{t+1} = \text{AdaDelta}(\mathbf{w}_{t}, L, \Delta\mathbf{w})$
		\ENDFOR
		\ENDFOR
		\STATE $\mathbf{w} = \text{BestBLEU}([\mathcal{E}]_{1}^{m})$
	\end{algorithmic}

\end{algorithm}

The above instance aggregating method can be used in a MERT-like iterative tuning algorithm as shown in Algorithm~\ref{alg-arc}, which can be easily integrated into current open source systems. The two standard listwise losses can be easily optimized using gradient-based methods (Algorithm~\ref{alg-training}); both losses are convex, so convergence to a global optimum is guaranteed. The gradients of ListNET and ListMLE with respect to the parameters $\mathbf{w}$ for a single sentence are:
\begin{equation*}
\begin{split}
&\frac{\partial L_{\text{Net-T}}}{\partial \mathbf{w}} =
- \sum_{j=1}^{k}P'_{eval}{(\mathbf{e}_{j})} \Biggl(\frac{\partial s(\mathbf{e}_{j})}{\partial \mathbf{w}} \\
&\quad {} - \sum_{j=1}^{k} \frac{\exp s(\mathbf{e}_{j})} {\sum_{j'=1}^{k}\exp{s(\mathbf{e}_{j'})}}\frac{\partial s(\mathbf{e}_{j})}{\partial \mathbf{w}}\Biggr)  \\
\end{split}
\end{equation*}

\begin{equation*}
\begin{split}
&\frac{\partial L_{\text{MLE}}}{\partial\mathbf{w}} = -\sum_{j=1}^{k}\Biggl( \frac{\partial s({\pi_{eval}(j)})}{\partial \mathbf{w}} \\
&{}\quad - \sum_{t=j}^{k}  \frac {\exp s({\pi_{eval}(t)})}{\sum_{t'=j}^{k}\exp s({\pi_{eval}(t')})} \frac{\partial s({\pi_{eval}(t)})}{\partial \mathbf{w}} \Biggr)\\
\end{split}
\end{equation*}

For optimization, We use a mini-batch stochastic gradient descent (SGD) algorithm together with AdaDelta~\cite{zeiler2012} algorithm to adaptively set the learning rate.

\section{Top-Rank Enhanced Losses}\label{sec-enhance}

In evaluating an SMT system, one naturally cares much more about the top-ranked results than the lower-ranked results. Therefore, we think that getting the ranking right at the top of a list is more relevant for tuning. Therefore, we should pay more attention to the top-ranked translations instead of forcing the model to rank the entire list correctly.

{\bf Position-dependent Attention: } To do this, we assign a higher cost to ranking errors that occur at the top and a lower cost to errors at the bottom. To make the cost sensitive to position, we define it as:
\begin{equation}\label{eq-cost}
c(j) = \frac{k - j + 1}{\sum_{t =1}^{k}t}
\end{equation}
where $j$ is the position in the ranking and $k$ is the size of the list.

Based on this cost function, we propose simple top-rank enhanced listwise losses as extensions of both the ListNet loss and the ListMLE loss. The loss functions are defined as follows:

\begin{equation*}
L_{\text{MLE-TE}} = -\sum_{j = 1}^{k} c(j) \log \frac {\exp s({\pi_{eval}(j)})} {\sum_{t = j}^{k}\exp s({\pi_{eval}(t)})}
\end{equation*}

\begin{align*}
L_{\text{Net-TE}} &= -\sum_{\forall \pi \in \Omega_{k}} P''_{\mathit{eval}}(\pi) \sum_{j=1}^k c(j) \log q_j(\pi) \\
q_j(\pi) &= {\frac {\exp z({\pi(j)})} {\sum_{t = j}^{k}\exp z({\pi(t)})}}.
\end{align*}

Along similar lines, \newcite{Xia2008} also proposed a top-$n$ ranking method, which assumes that only the correct ranking of top-$n$ hypotheses is useful. Compared to our top-rank enhanced losses, it may be too harsh to discard information about the rest of the ordering altogether; our method retains the whole ordering but weights it by position.

\section{Experiments and Results}\label{sec-exp}

\subsection{Data and Preparation}
We conduct experiments on a large scale Chinese-English translation task. The parallel data comes from LDC corpora\footnote{The corpora include LDC2002E18, LDC2003E14, LDC2004E12, LDC2004T08, LDC2005T10 and LDC2007T09}, which consists of 8.2 million of sentence pairs. Monolingual data includes Xinhua portion of Gigaword corpus. We use NIST MT03 evaluation test data as the development set, MT02, MT04 and MT05 as the test set.

\begin{table}[ht]\footnotesize
	\centering

	\begin{tabular}{c|c|cc}
		\cline{1-3}
		Data  & Usage & Sents. &  \\ \cline{1-3}
		LDC  & TM train & 8,260,093 &  \\ \cline{1-3}
		Gigaword  & LM train& 14,684,074 &  \\ \cline{1-3}
		MT03 & train & 919 & \\ \cline{1-3}
		MT02   & test & 878 & \\
		MT04   & test & 1,788 & \\
		MT05   & test & 1,082& \\ \cline{1-3}
	\end{tabular}
	\caption{Experimental data and statistics.}
	\label{tb-data}
\end{table}

The Chinese side of the corpora is word segmented using ICTCLAS\footnote{\url{http://ictclas.nlpir.org/}}. Word alignments of the parallel data are learned by running GIZA++~\cite{Och:2003} in both directions and refined under the ``grow-diag-final-and'' method. We train a 5-gram language model on the monolingual data with Modified Kneser-Ney smoothing\cite{Chen1999}. Throughout the experiments, our translation system is an in-house implementation of the hierarchical phrase-based translation system~\cite{Chiang2005}. The translation quality is evaluated by 4-gram case-insensitive BLEU~\cite{Papineni2002}. Statistical significance testing between systems is conducted by bootstrap re-sampling implemented by \newcite{Clark2011}.

\subsection{Tuning Settings}
We build baselines for extended and sparse feature sets with two different tuning methods. First, we tune with PRO~\cite{Hopkins:2011}. As reported by \newcite{Cherry2012}, it's hard to find the setting that performs well in general. We use MegaM version~\cite{Daume04} with 30 iterations for basic feature set and 100 iterations for extended and sparse feature sets. Second, we run the k-best batch MIRA (KB-MIRA) which shows comparable results with online version of MIRA~\cite{Cherry2012,Green2013}. In our experiments, we run KB-MIRA with standard settings in Moses\footnote{\url{http://www.statmt.org/moses/}}.
For the basic feature set, the baseline is tuned with MERT~\cite{Och2003}.

For all our listwise tuning methods, we set batch size  to 10. In our experiments, we can't find a epoch size perform well in general, so we set epoch size to 100 for ListMLE with basic features, 200 for ListMLE with extended and sparse features, and 300 for ListNet. These values are set to achieve the best performance on the development set.

We set beam size to 20 throughout our experiments unless otherwise noted. Following \newcite{Clark2011}, we run the same training procedure 3 times and present the average results for stability. All tuning methods are executed for 40 iterations of the outer loop and returned the weights that achieve the best development BLEU scores. For all tuning methods on sparse feature set, we use the weight vector tuned by PRO on the extended feature set as initial weights.

\subsection{Experiments of Listwise Learning Framework}
We first investigate the effectiveness of our instance aggregating training procedure. The results are presented in Table~\ref{table-merge}. The table compare training with instance aggregating and $k$-best merging. As the result suggested, with the instance aggregating method, the performance improves on both listwise tuning approaches. For the rest of this paper, we use the instance aggregating as standard setting for listwise tuning approaches.

\begin{table}\footnotesize
	\centering

	\begin{tabular}{l|c|c|c|c}
		\hline
		\multicolumn{1}{c|}{\textbf{Methods}} &\textbf{MT02}  &\textbf{MT04}  &\textbf{MT05}  &\textbf{AVG} \\ \hline
		Net$_{m}$            &40.36            &38.30            &37.93            &38.86(+0.00)        \\
		ListNet              &40.75            &38.69            &38.31            &39.25(+0.39)    \\ \hline
 		MLE$_{m}$            &39.82            &37.88            &37.65            &38.45(+0.00)         \\
		ListMLE              &40.40            &38.21            &38.04            &38.88(+0.43)      \\\hline
	\end{tabular}
	\caption{The comparison of instances aggregating and $k$-best merging on the extended feature set.(Net$_{m}$ and MLE$_{m}$ denote ListNet and ListMLE with $k$-best merging respectively.)}
	\label{table-merge}
\end{table}

\begin{table*}[t]\footnotesize
	\centering
    \begin{tabular}{lllllllll}
	\hline
	\multicolumn{1}{l|}{\multirow{2}{*}{\textbf{Method}}} & \multicolumn{4}{c|}{\textbf{Extended Features}}                                                                                                                       & \multicolumn{4}{c}{\textbf{Sparse Features}}                                                                                                                         \\ \cline{2-9}
	\multicolumn{1}{l|}{}                                 & \multicolumn{1}{c|}{\textbf{MT02}}  & \multicolumn{1}{c|}{\textbf{MT04}}  & \multicolumn{1}{c|}{\textbf{MT05}}  & \multicolumn{1}{c|}{\textbf{AVG}}          & \multicolumn{1}{c|}{\textbf{MT02}}  & \multicolumn{1}{c|}{\textbf{MT04}}  & \multicolumn{1}{c|}{\textbf{MT05}}  & \multicolumn{1}{c}{\textbf{AVG}}          \\ \hline
	\multicolumn{1}{l|}{PRO}                              & \multicolumn{1}{c|}{40.30}          & \multicolumn{1}{c|}{38.12}          & \multicolumn{1}{c|}{37.69}          & \multicolumn{1}{c|}{38.70(+0.00)}          & \multicolumn{1}{c|}{40.63}          & \multicolumn{1}{c|}{38.46}          & \multicolumn{1}{c|}{38.24}          & \multicolumn{1}{c}{39.11(+0.00)}          \\ \hline
	\multicolumn{1}{l|}{KB-MIRA}                          & \multicolumn{1}{c|}{40.48}          & \multicolumn{1}{c|}{37.71}          & \multicolumn{1}{c|}{37.37}          & \multicolumn{1}{c|}{38.52(-0.18)}          & \multicolumn{1}{c|}{40.67}          & \multicolumn{1}{c|}{38.48}          & \multicolumn{1}{c|}{38.21}          & \multicolumn{1}{c}{39.12(+0.01)}          \\ \hline
	\multicolumn{1}{l|}{ListNet}                          & \multicolumn{1}{c|}{\textbf{40.75$^*$}} & \multicolumn{1}{c|}{\textbf{38.69$^+$}} & \multicolumn{1}{c|}{\textbf{38.31$^*$}} & \multicolumn{1}{c|}{\textbf{39.25(+0.55)}} & \multicolumn{1}{c|}{\textbf{40.91$^*$}} & \multicolumn{1}{c|}{\textbf{38.77$^*$}} & \multicolumn{1}{c|}{\textbf{38.42}} & \multicolumn{1}{c}{\textbf{39.37(+0.26)}} \\ \hline
	\multicolumn{1}{l|}{ListMLE}                          & \multicolumn{1}{c|}{40.40}          & \multicolumn{1}{c|}{38.21}          & \multicolumn{1}{c|}{38.04}          & \multicolumn{1}{c|}{38.88(+0.18)}          & \multicolumn{1}{c|}{40.63}          & \multicolumn{1}{c|}{38.68}          & \multicolumn{1}{c|}{38.24}          & \multicolumn{1}{c}{39.18(+0.07)}          \\ \hline
	\hline
	\multicolumn{1}{l|}{ListMLE-T5}                                 & \multicolumn{1}{l|}{41.02$^*$}               & \multicolumn{1}{l|}{38.84$^+$}               & \multicolumn{1}{l|}{38.79$^+$}               & \multicolumn{1}{l|}{39.55(+0.85)}                      & \multicolumn{1}{l|}{41.12$^*$}               & \multicolumn{1}{l|}{ 38.91$^*$}               & \multicolumn{1}{l|}{38.89$^*$}               & \multicolumn{1}{l}{39.64(+0.53)}                      \\ \hline
	\multicolumn{1}{l|}{ListMLE-TE}                                 & \multicolumn{1}{l|}{\textbf{41.15$^+$}}               & \multicolumn{1}{l|}{\textbf{39.01$^+$}}               & \multicolumn{1}{l|}{\textbf{39.16$^+$}}               & \multicolumn{1}{l|}{\textbf{39.77(+1.07)}}                      & \multicolumn{1}{l|}{\textbf{41.25$^+$}}               & \multicolumn{1}{l|}{\textbf{39.00$^+$}}               & \multicolumn{1}{l|}{\textbf{39.27$^+$}}               & \multicolumn{1}{l}{\textbf{39.84(+0.73)}}                      \\ \hline
    \end{tabular}
    \caption{BLEU4 in percentage for comparing of baseline systems and systems with listwise losses. $^+$, $^*$ marks results that are significant better than the baseline system with $p<0.01$ and $p<0.05$. (ListMLE-T5 and ListMLE-TE refer to top-5 LisMLE and our top-rank enhanced ListMLE, respectively.)}
    \label{table-listwise}
\end{table*}

\begin{figure}
	\centering
	\includegraphics[width=.45\textwidth]{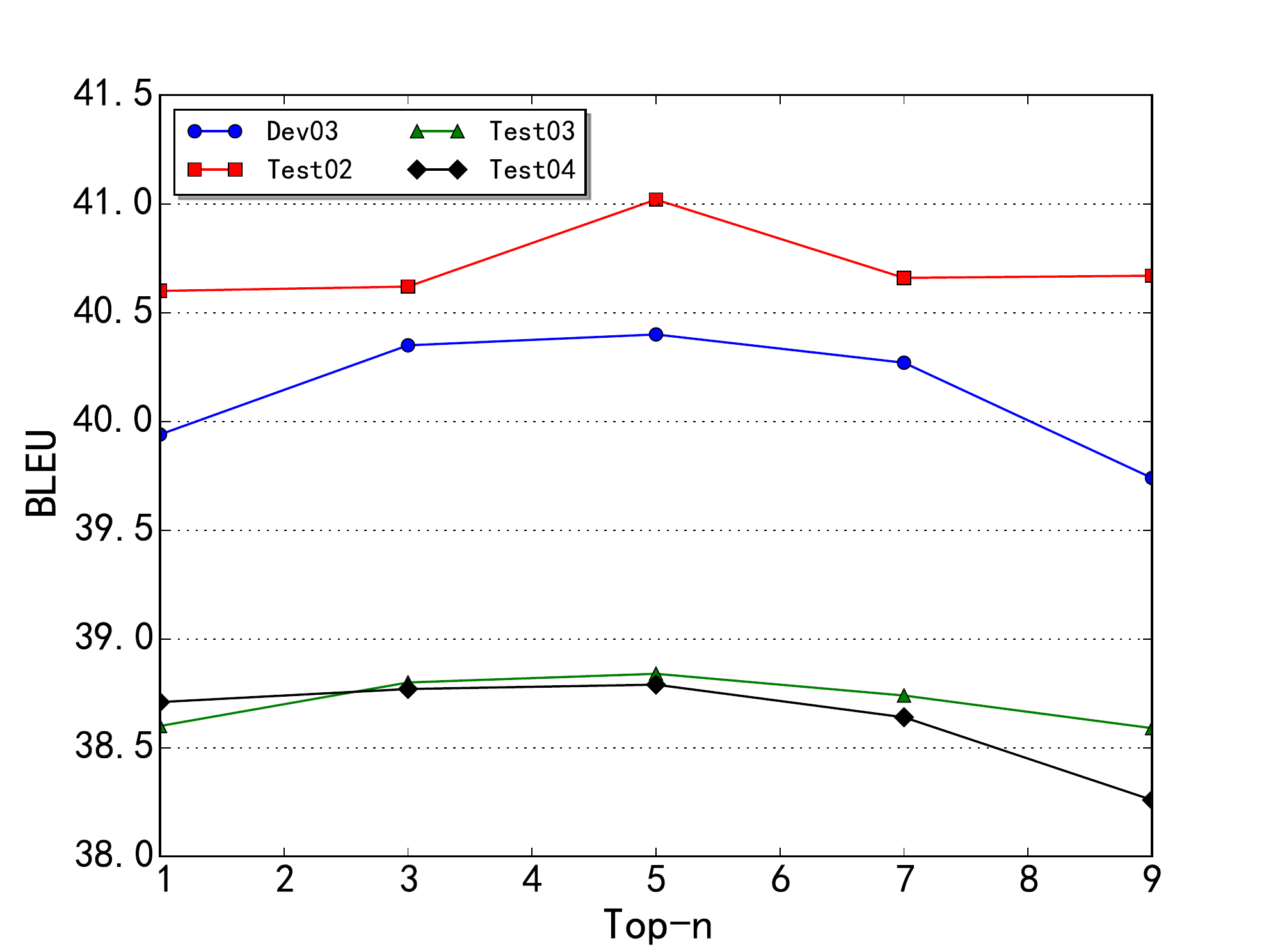}

    \caption{Effect of different $n$ for Top-$n$ ListMLE. We investigate the effect on the extended feature set.}
	
	\label{fig-top}
\end{figure}

To verify the performance of our proposed listwise learning framework, we first compare systems with standard listwise losses to the baseline systems. The first four rows in Table~\ref{table-listwise} show the results. ListNet can outperform PRO by 0.55 BLEU score and 0.26 BLEU score on extended feature set and sparse feature set, respectively. Its main reason is that our listwise methods can obtain structured order information when we take complete translation list as instance. 

We also observe that ListMLE can only get a modest performance compare to ListNet. We think the objective function of standard ListMLE which forces the whole list ranking in a correct order is too hard. ListNet mainly benefits from its top one permutation probability which only concerns the permutation with the best object ranked first.

\begin{figure*}[!t]
	
	\centering
	\begin{tabular}{@{}cc@{}}
		\subfloat[]{%
			\includegraphics[width=3in]{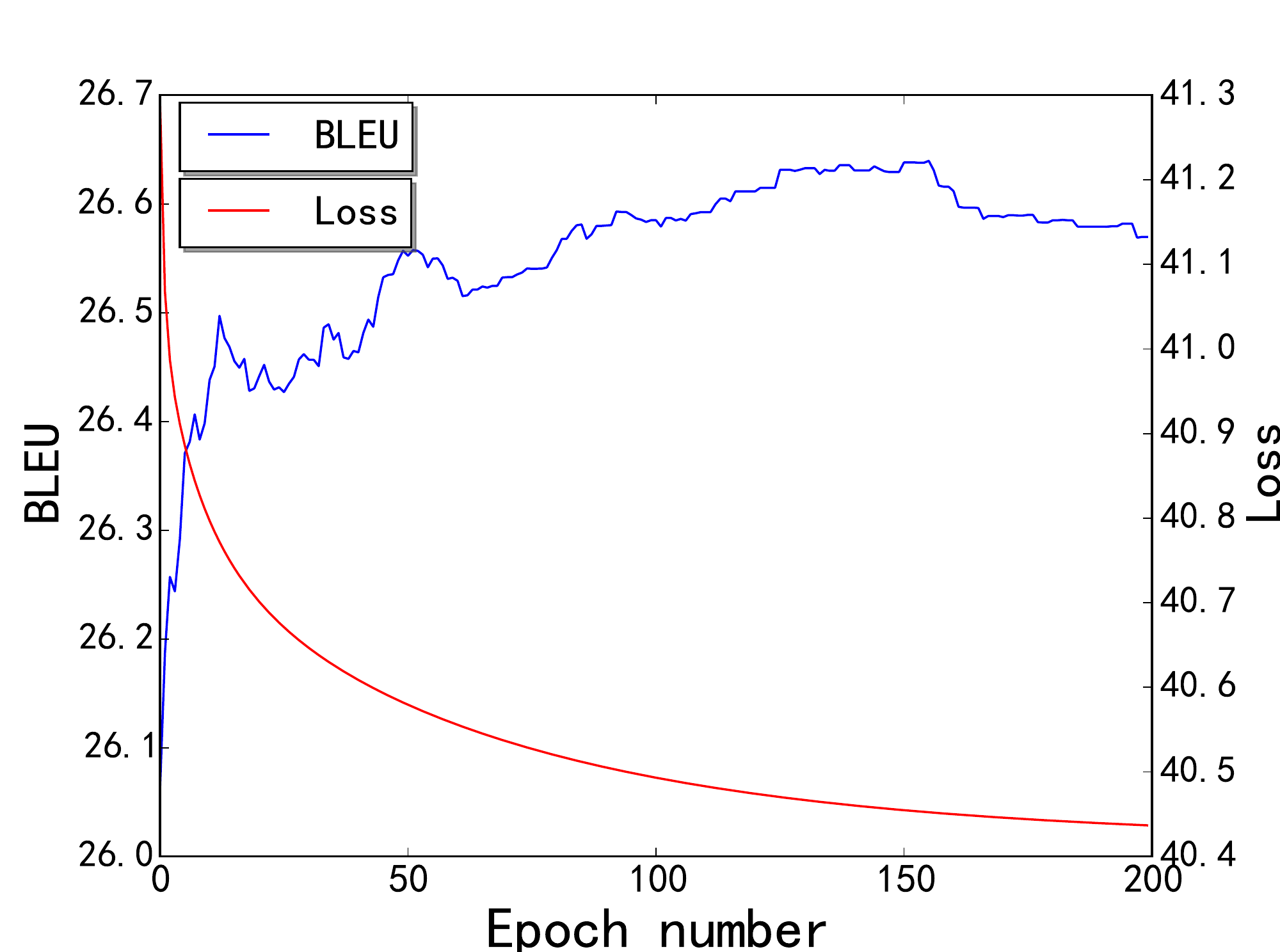}} &
		
		\subfloat[]{%
			\includegraphics[width=3in]{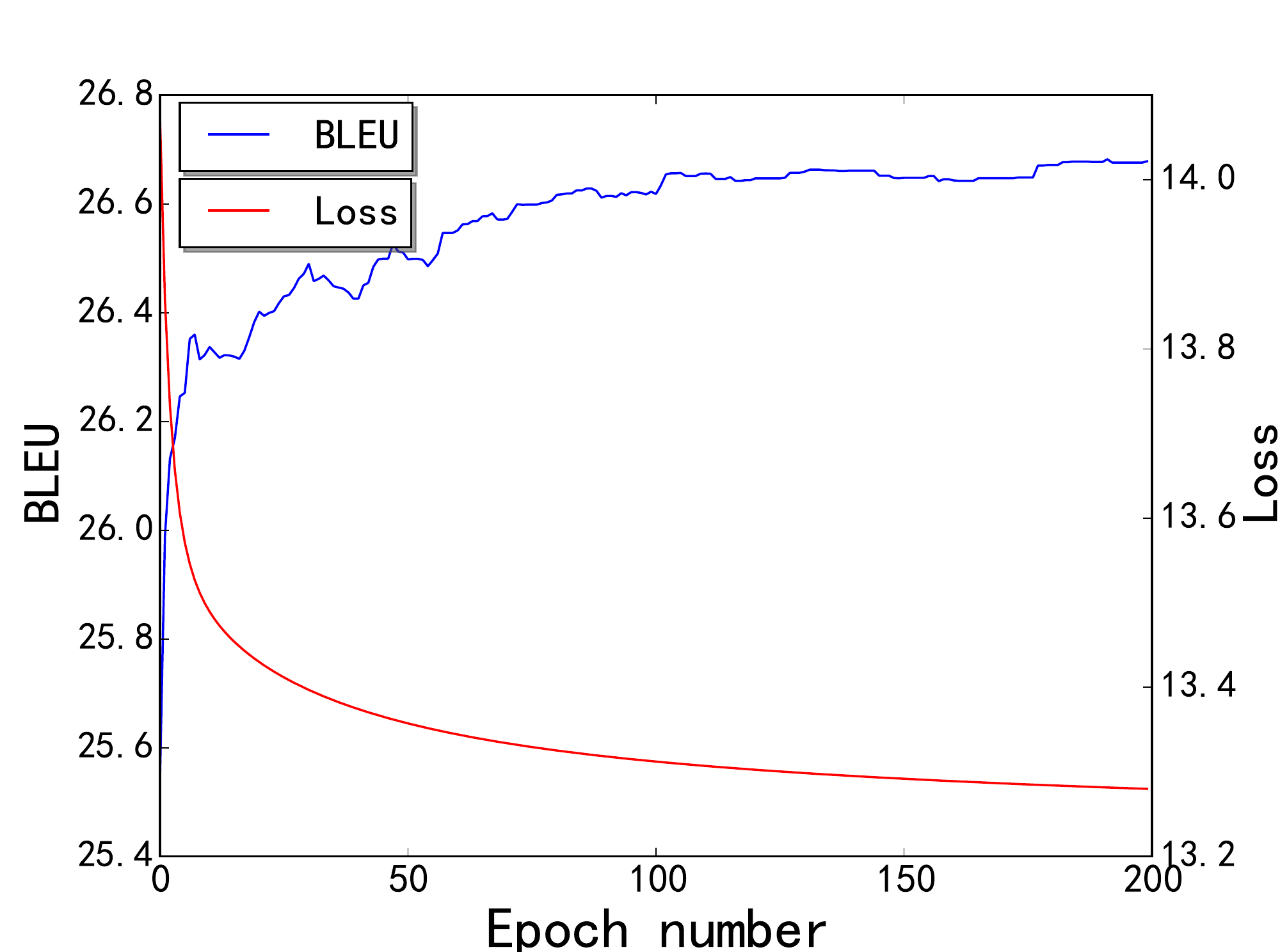}}\\
	\end{tabular}
	\caption{Listwise losses v.s. BLEU in (a) top-5 ListMLE and (b) top-rank enhanced ListMLE}
	
	\label{fig_losses}
	
\end{figure*}

\subsection{Effect of Top-rank Enhanced Losses}

To verify our assumption that the correct rank in the top portion of a list is more informative, we conduct this set of experiments. Figure~\ref{fig-top} shows the results of top-$n$ ListMLE with different $n$. Compared to ListMLE in Table~\ref{table-merge}, we find top-$n$ ListMLE can make significant improvements, which means that the top rank is more important. We can observe an improvement in all test sets when we set $n$ from 1 to 5, but when we further increase $n$, the results dropped. This situation indicates that the correct ranking at the top of the list is more informative and forcing the model to rank the bottom correctly as important as the top will sacrifice the ability to guide better search.

In Table~\ref{table-listwise}, top-5 ListMLE which only aims to rank the top five translations correctly can outperform the baseline and standard ListMLE. With our position-dependent attention, the top-rank enhanced ListMLE can make further improvement over the baseline system(+1.07 and +0.73 on extended and sparse feature sets, respectively.) and achieves the best performance. 

The top-$n$ loss might be too loose as an approximation of the measure of BLEU. Compared to top-$n$ ListMLE, our top-rank enhanced ListMLE can further utilize the different portions of the list by different weights. To verify the claim, we further examined the learning processes of the two losses. For simplicity, the experiment is conducted on a translation list generated by random parameters. The results are shown in Figure~\ref{fig_losses}. We can see that our top-rank enhanced loss almost completely inversely correlates with BLEU after iteration 70. In contrast, after iteration 150, although top-5 loss is still decreasing, BLEU starts to drop.

\begin{table}\scriptsize
	\centering
    \begin{tabular}{l|c|c|c|c}
		\hline
		\multicolumn{1}{c|}{\textbf{Methods}}  &\textbf{MT02}  &\textbf{MT04}  &\textbf{MT05}  &\textbf{AVG} \\ \hline
		PRO                & 40.90           & 38.84           & 38.64           & 39.64(+0.00)        \\
		KB-MIRA            & 41.09           & 38.49           & 38.62           & 39.40(-0.06)        \\
		ListNet            & 41.49$^+$       & 39.25$^*$       & 39.17$^*$       & 39.97(+0.51)        \\
		ListMLE-T5       & 41.26$^*$        & 39.63$^+$      & 39.32$^*$           & 40.07(+0.61)         \\
		ListMLE-TE      & \textbf{41.85$^+$}           & \textbf{39.96$^+$}          & \textbf{39.88$^+$}           & \textbf{40.56(+1.10)}        \\ \hline
	\end{tabular}
	\caption{Comparison of baselines and listwise approaches with a larger k-best list on extended feature set.}
	\label{table-kbest}
\end{table}

Due to the high computation cost of ListNet, we only perform the top-rank enhanced ListMLE in this paper. Our preliminary experiments indicate that the performance of ListNet can be further improved with a top-2 loss. We think our top-rank enhanced method is also useful for ListNet, but due to its computational demands it needs to be further investigated.

\subsection{Impact of the Size of Candidate Lists}\label{ssec-kbest}
Our listwise tuning methods directly model the order of the translation list, it is clear that the choice of the translation list size $k$ has an impact on our methods. A larger candidate list size may result in the availability of more information during tuning. In order to verify our tuning methods' capability of handling the larger translation list, we increase $k$ from 20 to 100. The comparison results are shown in Table~\ref{table-kbest}. With a larger size $k$, our tuning methods also perform better than baselines. For ListNet and top-5 ListMLE, we observe that the improvements over baseline is smaller than size 20. This results show that the order information loss caused by directly drop the bottom is aggravated with larger list size. However, our top-rank enhanced method still get a slight better result than size 20 and significant improvement over baseline by 1.1 BLEU score. This indicate that our top-rank enhanced method is more stable and can still effectively exploit the larger size translation list.

\subsection{Performance on Basic Feature Set}
Since the effectiveness of high dimensional feature set, recent work pays more attention to this scenario. Although previous discriminative tuning methods can effectively handle high dimensional feature set, MERT is still the dominant tuning method for basic features. Here, we investigate our top-rank enhanced tuning methods' capability of handling basic feature set. Table~\ref{table-basic} summarizes the comparison results. Firstly, we observe that ListNet and ListMLE can perform comparable with MERT. With our top-ranked enhanced method, we can get a better performance than MERT by 0.25 BLEU score. These results show that our top-ranked enhanced tuning method can learn more informations of translation list even with a basic feature set.

\begin{table} \scriptsize
	\centering
   \begin{tabular}{l|c|c|c|c}
		\hline
		\multicolumn{1}{c|}{\textbf{Methods}} &\textbf{MT02}  &\textbf{MT04}  &\textbf{MT05}  &\textbf{AVG} \\ \hline
		MERT              &37.72            &37.13            &36.77            &37.21(+0.00)        \\
		PRO               & 37.85           & 37.21           & 36.68           & 37.24(+0.03)        \\
		KB-MIRA           & 37.97           & 37.28           & 36.58           & 37.28(+0.07)       \\
		ListNet           & 37.71           & 37.47$^*$           & 36.78           & 37.32(+0.11)         \\
		ListMLE           &37.54             &\textbf{37.54}           &36.65             &37.24(+0.03)          \\
		ListMLE-T5      & 37.90           & 37.32           & 36.84           & 37.35(+0.14)         \\
		ListMLE-TE     & \textbf{38.03}           & 37.49$^*$         & \textbf{36.85}           & \textbf{37.46(+0.25)}        \\ \hline
	\end{tabular}
	\caption{Comparison of baseline and liswise approaches on basic feature set.}
	\label{table-basic}
\end{table}

\section{Related Work}
The ranking problem is well studied in IR community. There are many methods been proposed, including pointwise~\cite{Nallapati:2004}, pairwise~\cite{herbrich1999,Burges:2005} and listwise~\cite{Cao2007,Xia2008} algorithms. Experiment results show that listwise methods deliver better performance than pointwise and pairwise methods in general~\cite{Liu2010}.

Most NLP researches take ranking as an extra step after searching from its output space~\cite{Charniak:2005,Collins:2005,Duh:2008}. In SMT research, listwise approaches also have been employed for the reranking tasks. For example, \newcite{Li2013} utilized two listwise approaches to rerank the translation outputs and achieved the best segment-level correlation with human judgments. \newcite{Niehues2015} employed ListNet to rescore the $k$-best translations, 
which significantly outperforms MERT, KB-MIRA and PRO. \newcite{zhang2016} viewed the log-linear model as a pointwise ranking function and shifted it to listwise ranking function by introducing listwise features and outperformed the log-linear model. Compared to these efforts, our method takes a further step by integrating listwise ranking methods into the iterative training.

There are also some researches use ranking methods for tuning to guide better search. In SMT, previous attempts on using ranking as a tuning methods usually perform pairwise comparisons on a subset of translation pairs~\cite{Chiang2008,Hopkins:2011,Watanabe2012,Bazrafshan:2012,guzman-nakov-vogel:2015:CoNLL}. \newcite{Dreyer2015} even took all translation pairs of the $k$-best list as training instances, which only obtained a comparable result with PRO and the implementation is more complicate. In this paper, we model the entire list as a whole unit, and propose training objectives that are sensitive to different parts of the list.

\section{Conclusion}
In this paper, we propose a listwise learning framework for statistical machine translation. In order to adapt listwise approaches, we use an iterative training framework in which instances from different iterations are aggregated into the training set.
To emphasize the top order of the list, we further propose top-rank enhanced listwise learning losses.
Compared to previous efforts in SMT tuning, our method directly models the order information of the complete translation list. Experiments show our method could lead to significant improvements of translation quality in different feature sets and beam size.

Our current work focuses on the traditional SMT task. For future work, it will be interesting to integrate our methods to modern neural machine translation systems or other structure prediction problems. It may also be interesting to explore more methods on listwise tuning framework, such as investigating different methods to enhance top order of translation list directly w.r.t a given evaluation metric.

\section*{Acknowledgments}
The authors would like to thank the anonymous reviewers for their valuable comments. This work is supported by the
National Science Foundation of China (No. 61672277, 61300158 and 61472183). Part of Huadong Chen's contribution was made while visiting University of Notre Dame. His visit was supported by the joint PhD program of China Scholarship Council.

\bibliography{acl2017}
\bibliographystyle{acl_natbib}







\end{document}